\documentclass{article}

\usepackage[preprint]{neurips_2025} %[preprint]

\usepackage[utf8]{inputenc} % allow utf-8 input
\usepackage[T1]{fontenc}    % use 8-bit T1 fonts
\usepackage{hyperref}       % hyperlinks
\usepackage{url}            % simple URL typesetting
\usepackage{booktabs}       % professional-quality tables
\usepackage{amsfonts}       % blackboard math symbols
\usepackage{nicefrac}       % compact symbols for 1/2, etc.
\usepackage{microtype}      % microtypography
\usepackage{xcolor}         % colors
\usepackage{graphicx}
\usepackage{float}
\title{Toward Super Agent System with Hybrid AI Routers}
\author{%
  Yuhang Yao, Haixin Wang, Yibo Chen, Jiawen Wang, Min Chang Jordan Ren, \\ \textbf{Bosheng Ding, Salman Avestimehr, Chaoyang He}\\
  TensorOpera, Inc.\\
  \texttt{yuhang@tensoropera.com} \\
}

\begin{document}

\maketitle

\begin{abstract}
AI Agents powered by Large Language Models are transforming the world through enormous applications. A super agent has the potential to fulfill diverse user needs, such as summarization, coding, and research, by accurately understanding user intent and leveraging the appropriate tools to solve tasks. However, to make such an agent viable for real-world deployment and accessible at scale, significant optimizations are required to ensure high efficiency and low cost. This position paper presents a design of the \textit{Super Agent System} powered by the hybrid AI routers. Upon receiving a user prompt, the system first detects the intent of the user, then routes the request to specialized task agents with the necessary tools or automatically generates agentic workflows. In practice, most applications directly serve as AI assistants on edge devices such as phones and robots. As different language models vary in capability and cloud-based models often entail high computational costs, latency, and privacy concerns, we then explore the hybrid mode where the router dynamically selects between local and cloud models based on task complexity. Finally, we introduce the blueprint of an on-device super agent enhanced with cloud. With advances in multi-modality models and edge hardware, we envision that most computations can be handled locally, with cloud collaboration only as needed. Such architecture paves the way for super agents to be seamlessly integrated into everyday life in the near future.
\end{abstract}

\section{Introduction}
Artificial Intelligence Agents are rapidly evolving and are anticipated to profoundly reshape everyday human life within a few years~\cite{huang2024understanding,han2024llm}. Such sophisticated AI agents, often termed super agents, are designed to manage diverse tasks, ranging from routine actions such as sending emails to complex undertakings like software development and scientific research. The primary strength of a super agent is accurately interpreting human intent and effectively leveraging various tools and resources to address specific user needs~\cite{wei2025plangenllms}. As shown in Figure~\ref{fig:agent_system} (left), users experience a super agent through a simple interface: they ask questions and receive answers, while the underlying mechanisms remain hidden as a black box.

However, scaling such advanced agents to handle diverse user demands, especially at the level of serving billions of users concurrently, introduces significant challenges, primarily related to system reliability and performance under high request volumes~\cite{qian2024scaling}. For instance, popular chat services like ChatGPT and DeepSeek are prone to multi-hour outages~\cite{mather2025chatgpt} and often impose strict rate limits on tasks such as image and video generation~\cite{openai2025rate}. As agent complexity and real-world applications grow, these disruptions can lead to increasingly severe consequences.

To overcome these challenges and enable the robust deployment of super agents, \textbf{this position paper introduces a design of the \textit{Super Agent System} powered by hybrid AI routers}, an innovative AI agent-serving architecture illustrated in Figure~\ref{fig:agent_system}, which is built with four core components:

\begin{figure}[ht]
    \centering
    \includegraphics[width=0.98\textwidth]{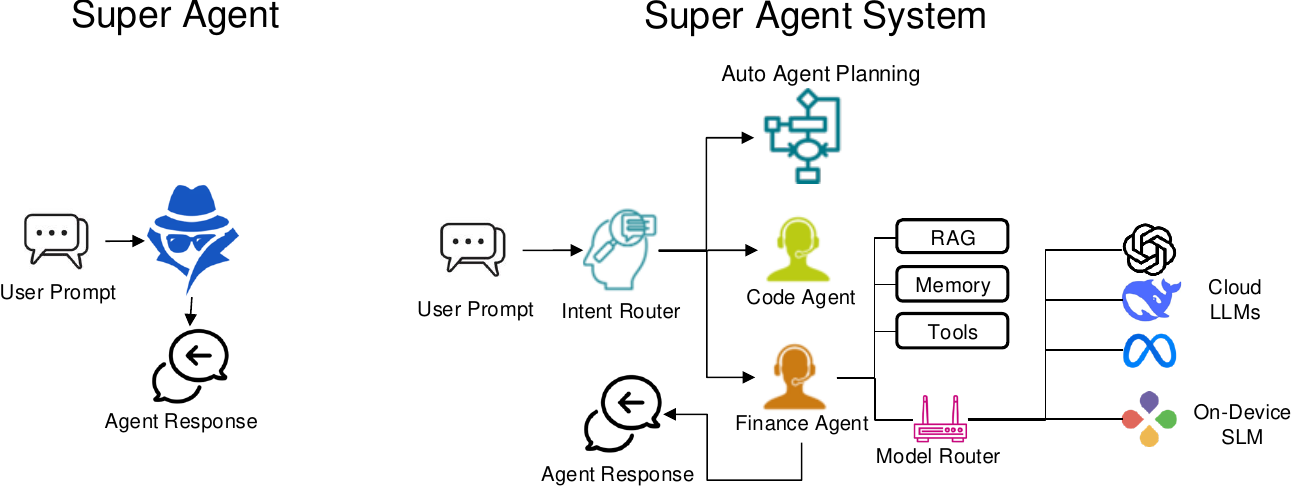}
    \caption{Overview of the \textit{Super Agent System}. The left figure illustrates the user-facing interface as a simple question-answer exchange. The right figure reveals the backend system architecture, including the intent router, task agents, model router, and hybrid edge-cloud language models. } %Together, these components enable efficient intent recognition, dynamic task execution, and scalable, cost-effective response generation.
    \label{fig:agent_system}
\end{figure}

\begin{itemize}
    \item \textbf{Intent Router and Planner}: Identifies the user's intent from prompts and routes the request to the appropriate agent, optimizing for efficiency and cost. For complex tasks involving multiple agents, the planner automatically generates a coordinated agentic workflow.
    
    \item \textbf{Task Agents with RAG, Memory, and Tools}: Specialized agents that execute identified tasks by leveraging Retrieval-Augmented Generation (RAG), shared memory, and external tools to improve task accuracy and efficiency.
    
    \item \textbf{Model Router}: Dynamically selects the most suitable language model based on task complexity. For example, directing the Olympic math problems to large reasoning models, daily coding tasks to medium-sized models, and text summarization to lightweight models.
    
    \item \textbf{On-Device SLM and Cloud LLMs}: Integrates a lightweight Small Language Model (SLM) running on edge devices with powerful cloud-based Large Language Models (LLMs) to balance latency, privacy, and computational cost.
\end{itemize}

By leveraging multiple specialized task agents along with automated agent planning, the system maximizes its ability to address a wide range of user demands with precision and efficiency. The intent router accurately directs each prompt to the most appropriate agent, while the model router dynamically selects the most suitable language model based on complexity and resource requirements across the edge-cloud language models. Together, these components ensure the \textit{Super Agent System} delivers high performance, responsiveness, and cost-effective services at scale.

In this paper, we first review the relative work in Section~\ref{sec:related_work}, then present the design principles of the proposed super agent serving architecture and provide detailed descriptions and rationale for each system component in Section~\ref{sec:design}. In Section~\ref{sec:on-device-super}, we introduce the blueprint of an on-device super agent that collaborates with cloud-based models and tools to optimize latency, accuracy, and cost. We conclude with a discussion of future research directions in Section~\ref{sec:conclustion_future_work}. %Furthermore, to facilitate practical deployments, we implement an end-to-end serving system prototype. Our experimental evaluations demonstrate the overall efficiency of our architecture and thoroughly explore the performance trade-offs involved in optimizing each component.

\section{Related Work}\label{sec:related_work}

\subsection{Intent Router}
Intent identification by LLMs~\cite{arora2024intent} forms the foundation of many agent-based applications.~\cite{bodonhelyi2024user} analyzes the quality of intent recognition and user satisfaction using intent-based prompt reformulations with GPT-3.5 Turbo and GPT-4 Turbo. IntentGPT~\cite{rodriguez2024intentgpt} introduces a training-free approach that effectively prompts LLMs such as GPT-4 to discover new intents with minimal labeled data.~\cite{sun2024large} proposes a semantic-level intent identification paradigm that improves recommendation quality. ~\cite{shah2023using} presents a human-in-the-loop pipeline for intent data generation.  LANID~\cite{fan2025lanid} and FSD-LLM~\cite{wang2024beyond} explore clustering-based techniques to discover novel user intents with LLM assistance.

\subsection{Planning}
Task planning is a core component of intelligent systems~\cite{huang2024understanding, wei2025plangenllms}. LLM-Planner~\cite{song2023llm} and LLM-DP~\cite{dagan2023dynamic} leverage few-shot prompting for task planning in embodied agents. LLM-Assist~\cite{sharan2023llm} combines rule-based and LLM-based planners for autonomous driving scenarios.~\cite{kambhampati2024position} introduces the LLM-Modulo framework, which integrates LLMs with external verifiers for robust bidirectional planning. ISR-LLM~\cite{zhou2024isr} enhances planning performance via iterative self-refinement. AutoHD~\cite{ling2025complex} enables LLMs to generate heuristic functions for improved inference-time search.~\cite{li2025parallelized} proposes a dual-threaded, interruptible framework for parallelized planning and acting in dynamic multi-agent environments. ~\cite{nie2025resource} translates a task into a program and learns a policy that selects foundation model backends for each program module.

\subsection{Model Router}
Model routing plays a critical role in balancing model accuracy with serving cost. Hybrid LLM~\cite{ding2024hybrid} employs a difficulty-aware router that selects between small and large models. TO-Router~\cite{stripelis2024tensoropera} introduces a modular querying system that dynamically routes queries to high-performing experts. GraphRouter~\cite{feng2024graphrouter} proposes a graph-based routing strategy for LLM selection. RouterBench~\cite{hu2024routerbench} offers a benchmark suite for evaluating multi-LLM routing systems. RouterLLM~\cite{ong2024routellmlearningroutellms} provides a training framework to learn efficient model selectors that balance performance and cost. P2L~\cite{frick2025prompt} estimates LLM performance by mapping prompts to a leaderboard score. Online Cascade Learning~\cite{nie2024online} inferences starting with lower-capacity models and ending with a powerful LLM. Nvidia's LLM-Router~\cite{nvidia_llm_router} routes prompts based on task type and complexity using pre-trained classifiers.

Collectively, these research efforts form the foundational components for building a unified and scalable \textit{Super Agent System}.

\section{Super Agent System Design}\label{sec:design}
In this section, we begin by outlining the key design principles guiding the development of the \textit{Super Agent System}. We then present a detailed breakdown of the system architecture and explain the rationale behind each of its core components.
\subsection{Design Principles}
We first outline the design principles that prioritize both user-centric interaction and system-level efficiency.

\textbf{User simplicity with comprehensive functionality}: The user interface is designed to be minimal, a simple chat window, while supporting a wide range of user demands seamlessly.

\textbf{Accurate intent identification}: User requests must be precisely routed to the most appropriate agent, ensuring effective execution through proper alignment of capabilities and resources.

\textbf{Robust task agents}: Each specialized agent should possess strong task-solving capabilities, leveraging external tools and memory to handle complex requests reliably and efficiently.

\textbf{Efficient model routing for cost-effectiveness}: The system should incorporate a smart language model router that allocates tasks to the most suitable models, optimizing for both performance and resource utilization under high user demand.

We now present the detailed design of each core component within the \textit{Super Agent System}.

\subsection{Intent Router}

\begin{figure}[ht]
\centering
\includegraphics[width=0.7\textwidth]{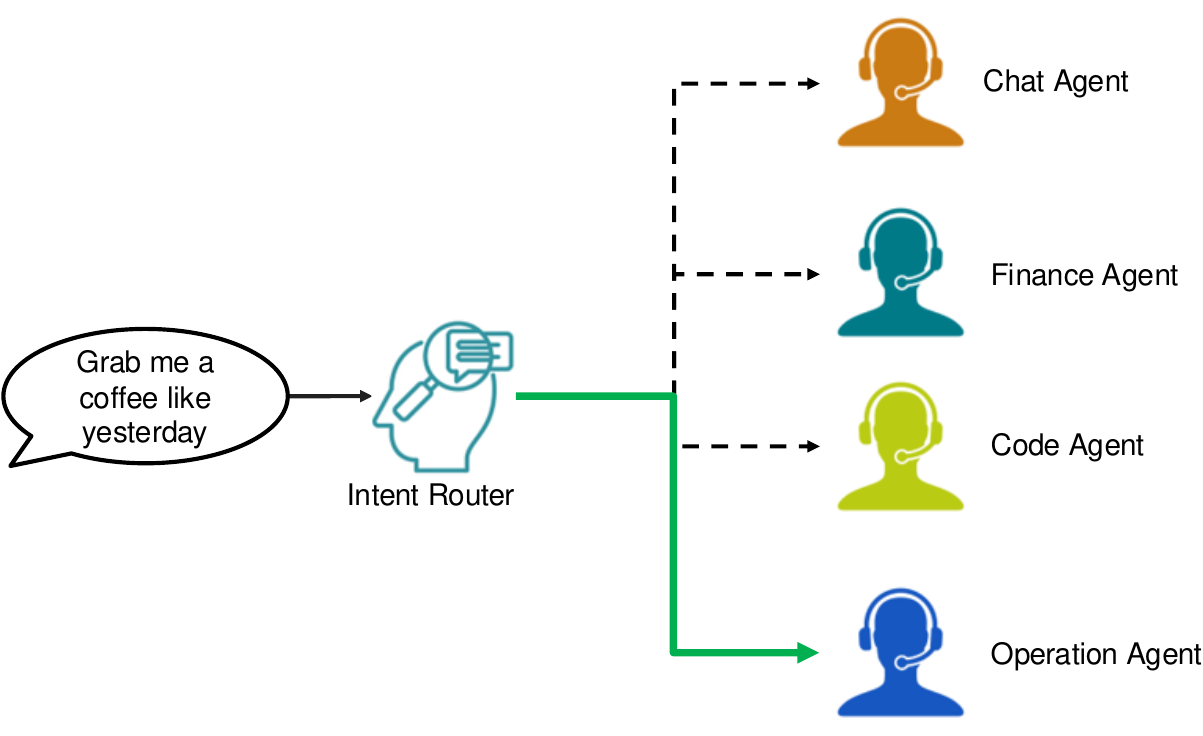}
\caption{Illustration of the Intent Identification Process. The intent router analyzes prompts and directs them to the corresponding task agents, such as chat, finance, code, or operation agents, based on the detected intent.}
\label{fig:intent_router}
\end{figure}

The intent router is responsible for interpreting user prompts and directing them to the appropriate downstream agents. By analyzing the semantics and context of each input, the router ensures that tasks are executed by the most relevant and capable agent. As illustrated in Figure~\ref{fig:intent_router}, for example, a prompt like "Grab me a coffee like yesterday" might be routed to the Operation Agent, which can leverage user history and external services to fulfill the request. This routing mechanism enables targeted, context-aware responses by dynamically invoking the most relevant agent or workflow.

\subsubsection{Intent as Function Call}
One approach to implementing the intent router is through prompt engineering, where the language model is instructed to choose from a predefined list of agents by returning the agent's name directly~\cite{arora2024intent}. However, this method relies heavily on precise prompt design and typically requires large, high-quality models to achieve reliable performance.

To address these limitations, we adopt a more structured method using the Function Call~\cite{erdogan2024tinyagent}, where each agent is represented as a callable function. This enables the model to return not only the agent name but also relevant arguments, such as the confidence level, in a standardized format. As a result, the routing process becomes more interpretable and extensible.

\begin{figure}[ht]
    \centering
    \includegraphics[width=0.7\textwidth]{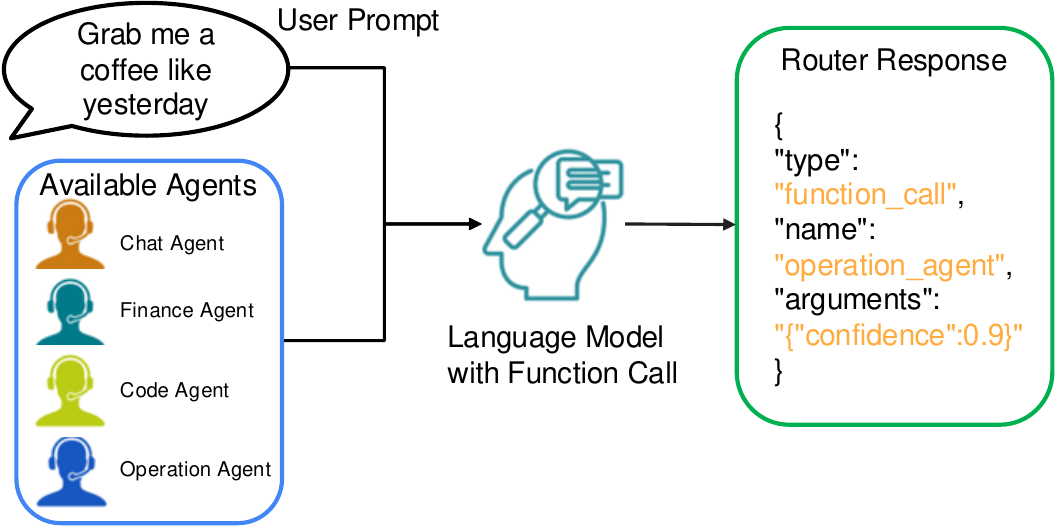}
    \caption{Classifying User Intent by Function Call. The model receives a prompt and a list of available agents, then outputs a structured function call with the selected agent and routing parameters.}
    \label{fig:function_call}
\end{figure}

Importantly, this approach can be implemented without fine-tuning by leveraging small language models~\cite{grattafiori2024llama, hu2024fox} that support function call and operate with low latency, making it suitable for real-time applications. As shown in Figure~\ref{fig:function_call}, the user prompt and list of available agents are passed into the model. The model responds with a function call to operation\_agent, along with a confidence score of 0.9. The system then routes the request to the corresponding agent for execution.

\subsubsection{Auto Agent Workflow Planning}

Automated planning involving single or multiple agents is an evolving area and remains in early stages for practical deployment~\cite{wei2025plangenllms}. With growing interest from research and industry, the ability to generate agentic workflows, where multiple agents and tools collaborate dynamically, is becoming increasingly feasible. Frameworks such as Google’s agent-to-agent (A2A) architecture~\cite{surapaneni2025a2a} exemplify this trend by enabling agents to delegate tasks and coordinate actions autonomously.

\begin{figure}[ht]
    \centering
    \includegraphics[width=0.8\linewidth]{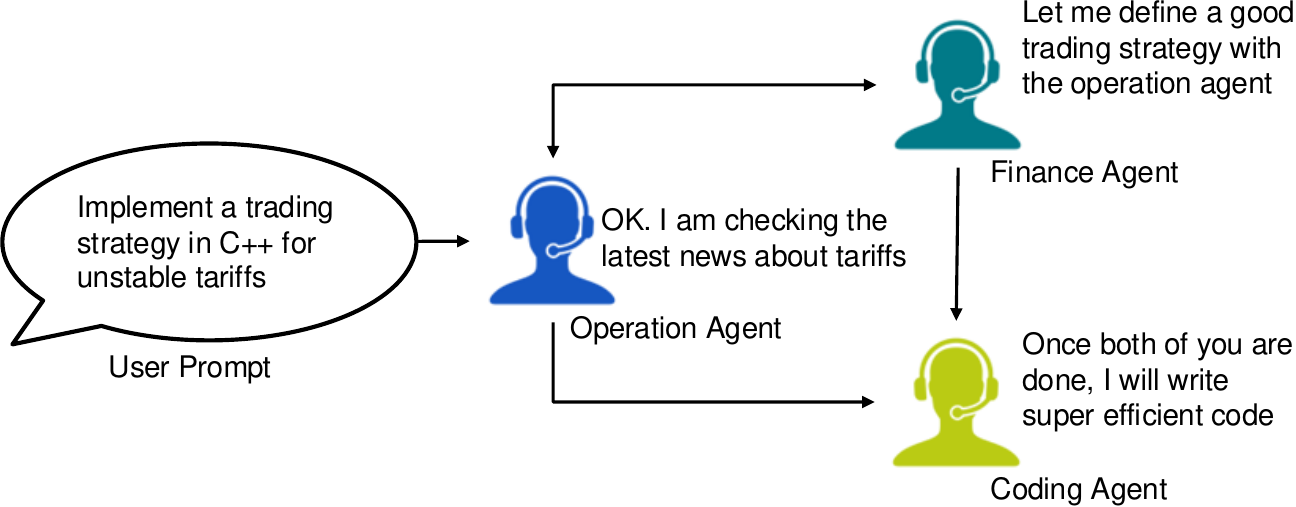}
    \caption{Auto Agent Workflow Planning. A single user prompt initiates a collaborative process across multiple agents, each handling a specialized aspect of the task, from context gathering to strategy design and code generation.}
    \label{fig:auto_agentic_planning}
\end{figure}

As illustrated in Figure~\ref{fig:auto_agentic_planning}, a user prompt such as "Implement a trading strategy in C++ for unstable tariffs" can trigger a multi-agent collaboration: the Operation Agent gathers up-to-date tariff information, the Finance Agent formulates a strategy, and the Coding Agent ultimately generates the corresponding C++ implementation. This type of automated, task-oriented planning significantly improves modularity, scalability, and adaptability in complex workflows.

\subsection{Task Agents}

Once the user intent is identified and the execution plan is established, the designated task agent, an individual agent with strong domain-specific capabilities, takes over. Task agents are responsible for executing downstream tasks and are typically organized into structured agentic workflows. These workflows benefit from platforms such as LangFlow~\cite{Langflow} and AutoGen~\cite{wu2023autogen}, which facilitate easy configuration and deployment of complex agent interactions.

\subsubsection{RAG, Memory, and Tool Use}
To perform tasks effectively, task agents rely on several key components: retrieval-augmented generation (RAG) for real-time access to relevant external or internal information; memory modules for maintaining user-specific context across sessions; and tool integration for interfacing with external services or APIs. Together, these elements enable personalized, context-aware responses and autonomous task execution.

\begin{figure}[ht]
    \centering
    \includegraphics[width=0.8\linewidth]{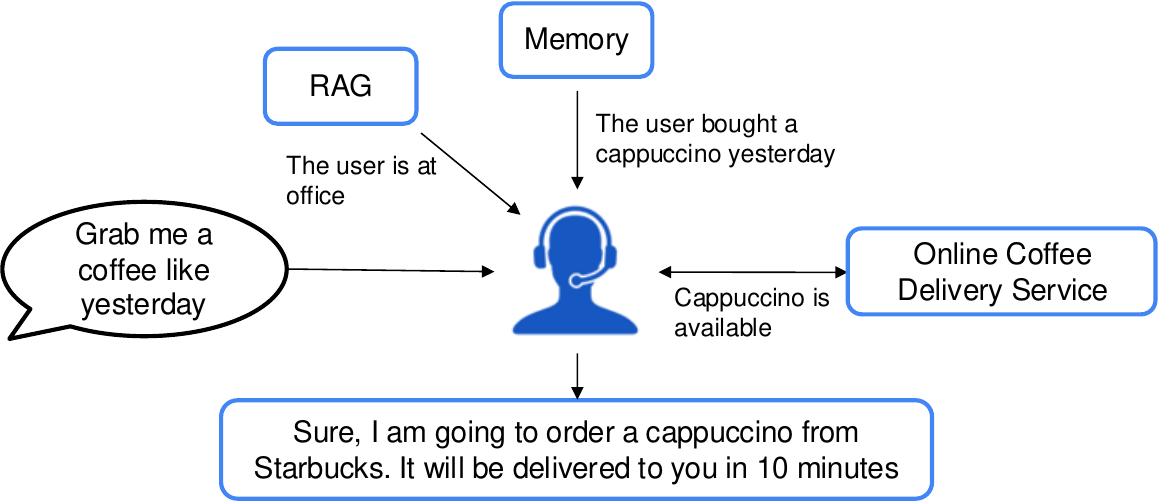}
    \caption{Example of a Task Agent. The agent retrieves context from memory and RAG, verifies item availability through external tools, and performs task execution.}
    \label{fig:task_agent_example}
\end{figure}

\begin{figure}[ht]
    \centering
    \includegraphics[width=0.8\linewidth]{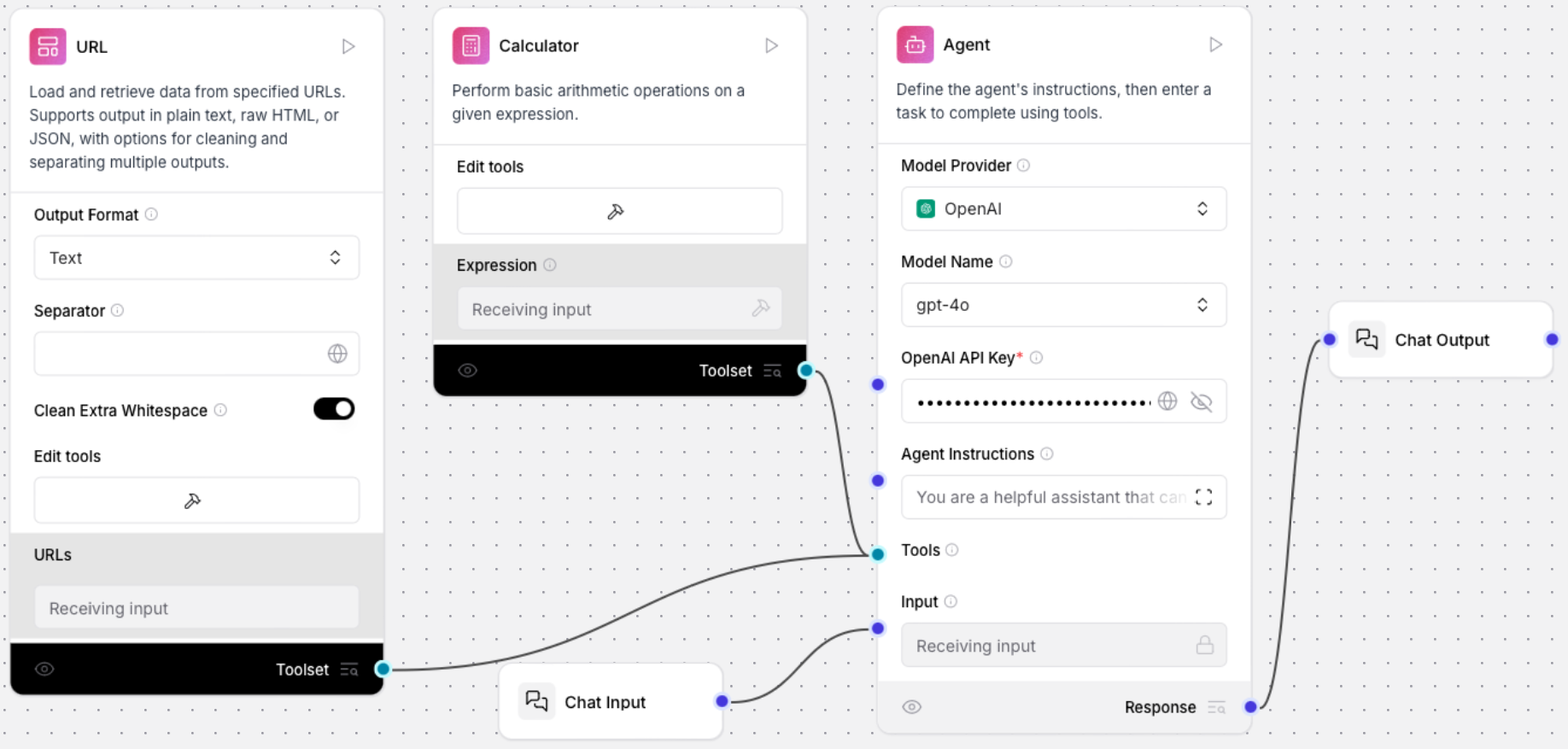}
    \caption{Agentic Workflow Implemented by LangFlow. The agent integrates tools such as website retrieval and a calculator, processes user input via a language model, and generates responses.}
    \label{fig:agentic_workflow}
\end{figure}

Figure~\ref{fig:task_agent_example} shows a practical example of such a task agent. In this scenario, a user request like "Grab me a coffee like yesterday" activates a task agent that retrieves contextual data from memory (e.g., the user ordered a cappuccino yesterday), gathers real-time delivery options using RAG and tool APIs, and executes the action by placing an order through an online coffee delivery service. The agent then generates a natural response confirming the completed action.

\subsubsection{Agentic Workflow Implementation}
The agentic workflow orchestrates a sequence of subtasks by integrating tools, memory modules, and external services to fulfill user requests in a contextualized and efficient manner. This modular structure enables agents to break down complex tasks into manageable components, delegate them to specialized tools, and produce accurate, timely responses.

Figure~\ref{fig:agentic_workflow} presents a real-world example of such a workflow using LangFlow~\cite{Langflow}. In this configuration, an agent powered by a language model (e.g., GPT-4o) orchestrates multiple tool calls, such as retrieving web data via a website parser and performing computations through a calculator, before generating a final response through the chat interface. This illustrates how agents can dynamically combine reasoning with tool use to complete multi-step tasks autonomously.

\subsection{Model Router}
Task agents rely heavily on large language models to perform tool calling, reasoning, and user interaction. To balance performance and cost, the model router dynamically selects from multiple available models based on the specific requirements of each task. This routing mechanism ensures that high-accuracy responses are delivered when needed, while also minimizing computational overhead where possible. We then introduce two routing modes that reflect common optimization objectives in real-world applications.
\subsubsection{Accuracy-Optimized Routing}
This mode prioritizes selecting the most capable LLMs for a given task, focusing on maximizing performance quality and precision. It is especially valuable for complex queries where response accuracy is critical (e.g., Prompt-to-Leaderboard~\cite{frick2025prompt}).

\subsubsection{Cost-Optimized Routing}
\begin{figure}[ht]
    \centering
    \includegraphics[width=0.48\linewidth]{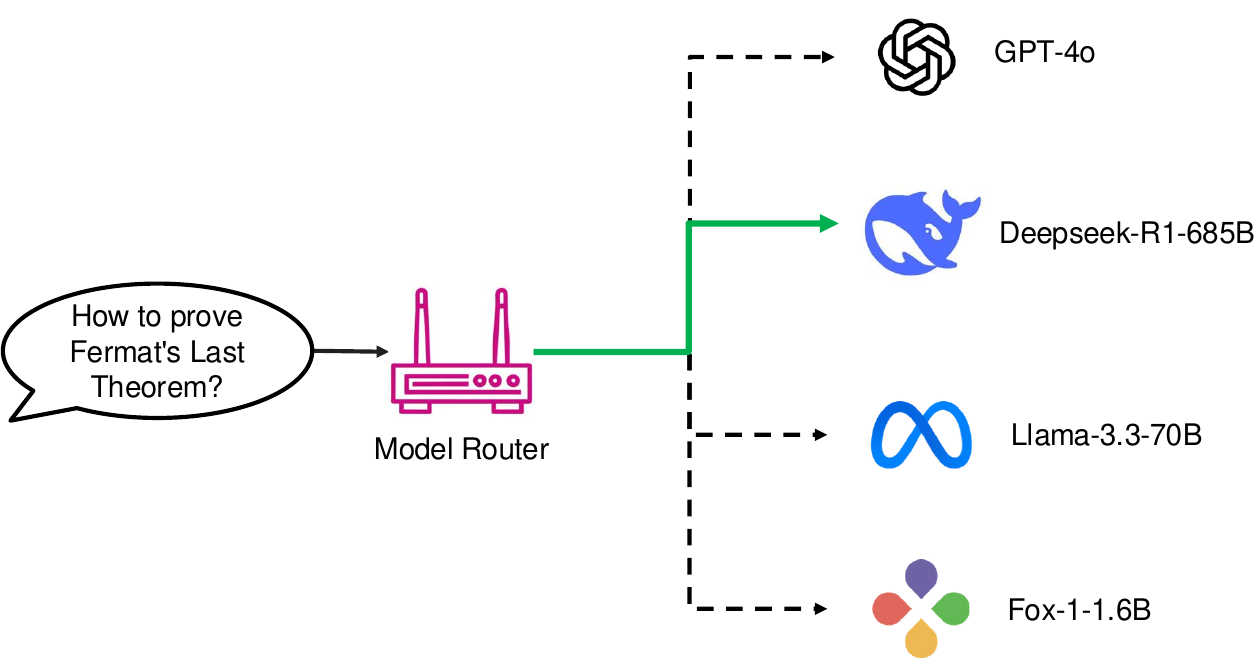}
    \hspace{2mm}
    \includegraphics[width=0.48\linewidth]{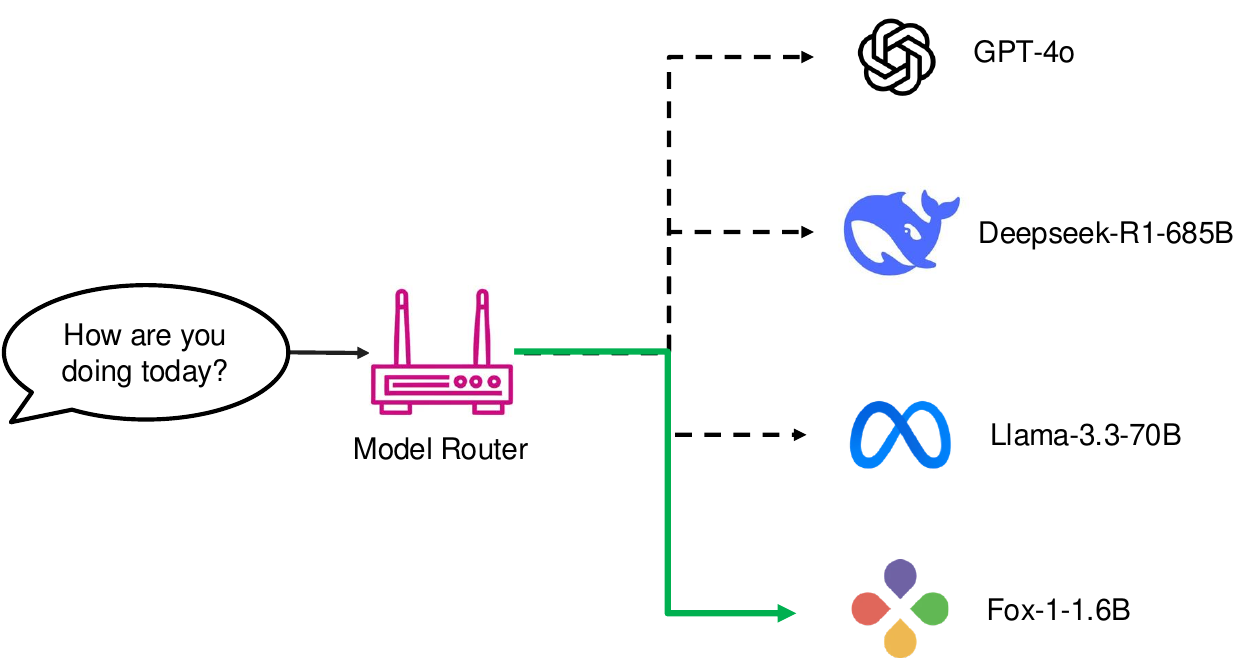}
    \caption{Model Router with Cost-Optimized Configurations. The router dynamically selects language models based on task complexity, balancing precision and efficiency.}
    \label{fig:llm_router}
\end{figure}

This mode aims to reduce computational costs by routing tasks to lightweight or less expensive models, while maintaining acceptable performance. Techniques such as TensorOpera Router~\cite{stripelis2024tensoropera} and RouteLLM~\cite{ong2024routellmlearningroutellms} exemplify this trade-off by dynamically adapting model selection based on task difficulty and resource availability. As shown in Figure~\ref{fig:llm_router}, the router routes complex mathematical queries to high-performance models, while assigning simpler conversational tasks to smaller, more cost-efficient models.

\subsection{Edge-Cloud Router}
In real-world applications, AI agents often operate on user devices such as smartphones and robots. Relying solely on cloud-based models introduces high latency, bandwidth consumption, and privacy concerns. Edge-cloud router optimizes performance by offloading most computations to an on-device small language model (SLM), enabling fast, privacy-preserving responses, while selectively routing complex or resource-intensive tasks to cloud-based large language models~\cite{stripelis2024tensoropera}.

\begin{figure}[ht]
    \centering
    \includegraphics[width=0.9\linewidth]{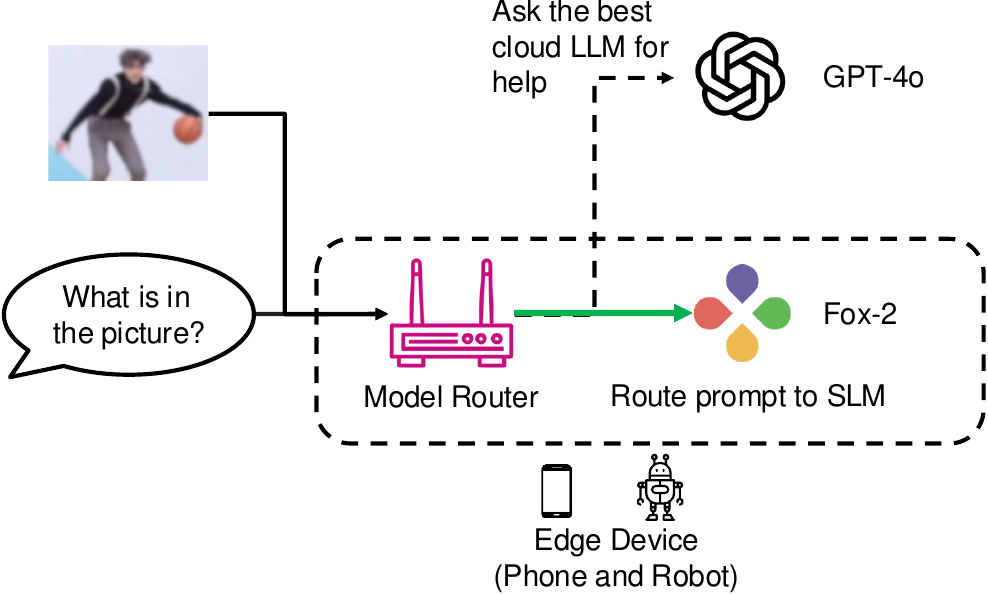}
    \caption{Edge-Cloud Router. A model router on the edge device first attempts to resolve user queries locally via an SLM. Only when necessary, it sends complex tasks to the best cloud-based LLM. }
    \label{fig:edge_cloud_router}
\end{figure}

As shown in Figure~\ref{fig:edge_cloud_router}, given an image recognition task, a model router embedded in the edge device first attempts to handle the user prompt using an efficient on-device small language model, such as Fox-2. If the local model is insufficient to handle the request, the router sends the request to a powerful cloud-based model like GPT-4o. This hybrid routing mechanism ensures an optimal balance of latency, cost, accuracy, and privacy.

\section{Blueprint: On-Device Super Agent Enhanced with Cloud}\label{sec:on-device-super}
With the increasing capabilities of edge hardware, a small language model (SLM) running directly on the device can serve as a super agent, capable of handling user requests with low latency, enhanced privacy, and minimal reliance on external infrastructure.

\begin{figure}[ht]
    \centering
    \includegraphics[width=0.97\linewidth]{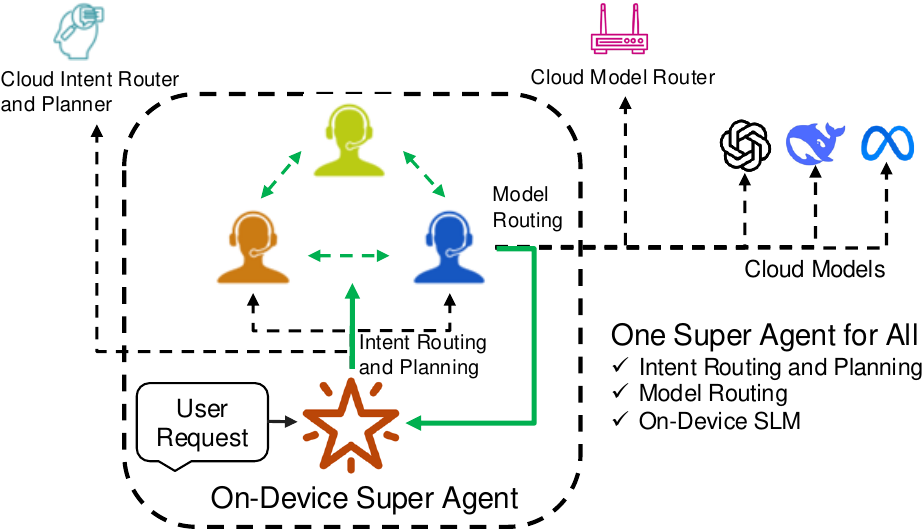}
    \caption{On-Device Super Agent. A unified agent on edge devices performs intent routing, planning, and model selection locally. For complex tasks, the system offloads execution to cloud-based routers and LLMs, enabling seamless collaboration between local and cloud resources.}
    \label{fig:on-device-super-agent}
\end{figure}

As shown in Figure~\ref{fig:on-device-super-agent}, the on-device super agent integrates intent routing, task planning, and model selection, enabling local execution for most tasks. When handling more complex queries or requiring high computational resources, the system can seamlessly delegate subtasks to cloud-based intent routers, model routers, or large language models.

This hybrid architecture also ensures robustness and continuity: in offline scenarios, the on-device agent operates independently; when connectivity is available, it leverages the cloud to enhance capability and scalability. This blueprint offers a flexible, privacy-preserving, and cost-efficient solution for deploying super agents across phones, robots, and other edge platforms.

%

% \section{Experiments}

% \subsection{End to End}
% System tracing and breakdown
% \subsection{Agent Router}
% Cloud only, edge only, and Edge-Cloud

% \subsection{Tool use and RAG}

% \subsection{Model Router}
% Cloud only, edge only, and Edge-Cloud

\section{Conclusion and Future Work}\label{sec:conclustion_future_work}
In this paper, we proposed the \textit{Super Agent System} with hybrid AI routers, a modular and scalable architecture for building intelligent agents capable of handling diverse user requests through dynamic intent routing, agent planning, and model selection across edge and cloud environments.

We finally identify several key directions for future research and system development.

\textbf{Benchmarking}: Establish benchmarks for the end-to-end system, evaluating each component independently to identify performance bottlenecks and areas for improvement.

\textbf{System Optimization}: Develop new algorithms to improve the overall efficiency, responsiveness, and scalability of the agent system, particularly under high user demand.

\textbf{Train Super Agent System}: In real-world deployments, user feedback can be leveraged to train reward models for evaluating agent performance. These insights can then be used to further fine-tune the system via supervised learning or reinforcement learning.

\textbf{Task Agent Optimization and Planning}: For specialized agents (e.g., for coding or finance), workflows can be refined either manually or through automated planning with large language models to enhance task execution quality and efficiency.

\bibliography{ref}
\bibliographystyle{ACM-Reference-Format}

\appendix

\end{document}